\documentclass[conference]{IEEEtran}
\usepackage{cite}
\usepackage{amsmath,amssymb,amsfonts}
\usepackage{algorithmic}
\usepackage{graphicx}
\usepackage{textcomp}
\usepackage{xcolor}
\usepackage{subcaption}
\usepackage{booktabs}
\usepackage{tabularx}
\usepackage{hyperref}
\newcolumntype{Y}{>{\centering\arraybackslash}X}

\def\BibTeX{{\rm B\kern-.05em{\sc i\kern-.025em b}\kern-.08em
    T\kern-.1667em\lower.7ex\hbox{E}\kern-.125emX}}
\begin{document}

\title{Data Augmentation Techniques for Cross-Domain WiFi CSI-based Human Activity Recognition
\thanks{Identify applicable funding agency here. If none, delete this.}
}

\author{\IEEEauthorblockN{Julian Strohmayer and Martin Kampel}
\IEEEauthorblockA{\textit{Computer Vision Lab, TU Wien} \\
Favoritenstr. 9/193-1, 1040 Vienna, Austria \\
\{julian.strohmayer, martin.kampel\}@tuwien.ac.at}
}

\maketitle

\begin{abstract}
The recognition of human activities based on WiFi Channel State Information (CSI) enables contactless and visual privacy-preserving sensing in indoor environments. However, poor model generalization, due to varying environmental conditions and sensing hardware, is a well-known problem in this space. To address this issue, in this work, data augmentation techniques commonly used in image-based learning are applied to WiFi CSI to investigate their effects on model generalization performance in cross-scenario and cross-system settings. In particular, we focus on the generalization between line-of-sight (LOS) and non-line-of-sight (NLOS) through-wall scenarios, as well as on the generalization between different antenna systems, which remains under-explored. We collect and make publicly available a dataset of CSI amplitude spectrograms of human activities. Utilizing this data, an ablation study is conducted in which activity recognition models based on the EfficientNetV2 architecture are trained, allowing us to assess the effects of each augmentation on model generalization performance. The gathered results show that specific combinations of simple data augmentation techniques applied to CSI amplitude data can significantly improve cross-scenario and cross-system generalization.
\end{abstract}

\begin{IEEEkeywords}
Data Augmentation, Model Generalization, Human Activity Recognition, WiFi, Channel State Information
\end{IEEEkeywords}

\section{Introduction}
\label{sec:introduction}
\begin{figure}[t!]
  \centering
  \includegraphics[width=\linewidth]{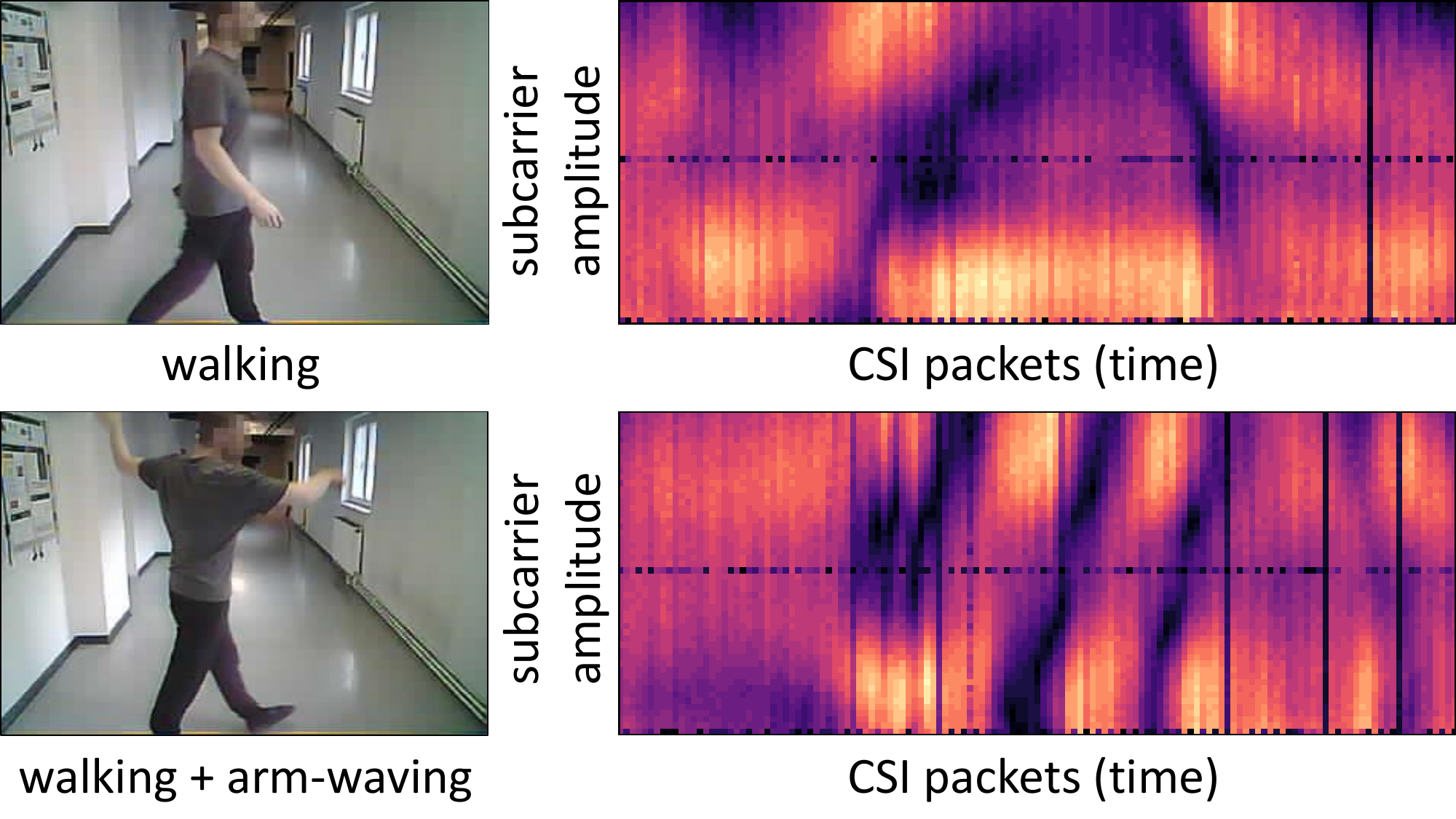}
  \vspace{-6mm}
  \caption{CSI spectrograms of a person walking and a person walking while simultaneously waving their arms, showing the characteristic patterns in WiFi subcarrier amplitudes caused by these activities.}
  \label{fig:activitiesCloseup}
  \vspace{-5mm}
\end{figure}

WiFi Channel State Information (CSI)-based Human Activity Recognition (HAR) has garnered significant attention due to its advantages over optical modalities, such as preserving visual privacy, being unobtrusive, and having wall-penetration capabilities that enable monitoring of large areas with a single system \cite{Strohmayer2023}. In indoor settings, the propagation paths of WiFi signals are influenced by both static and dynamic objects that constitute the environment \cite{Lee10185958}. Static objects, such as walls, floors, or furniture, primarily contribute to the background signal. On the other hand, the motion of dynamic objects, such as the human body, can significantly alter signal propagation paths within short periods of time. Consequently, characteristic patterns in CSI emerge, enabling the distinction between various activities, as illustrated in Figure \ref{fig:activitiesCloseup}. This forms the foundation of CSI-based HAR. However, a consequence of this inherent sensitivity to variations in the environment, sensing hardware, or the physiology of monitored individuals, all of which contribute to CSI characteristics, make model generalization to new domains an open problem \cite{Chen2023} often resulting in a decline in performance due to an underlying domain gap \cite{Qiao2020LearningTL}.

To address this problem, in this work, we investigate methods to enhance both cross-scenario and cross-system generalization performance of WiFi CSI-based HAR models. Cross-scenario generalization pertains to a model's ability to perform well on CSI from a scenario it was not trained on, without encountering samples from the target domain during training. Similarly, cross-system generalization involves a model's capability to perform well when tested on CSI from a different system than the one it was trained on. Developing methodologies to improve cross-scenario and cross-system generalization could enable knowledge transfer between domains and enhance a model's robustness against variations in environmental conditions and sensing hardware, frequently encountered in practical deployments. Furthermore, it could address the limited availability of data in the field of WiFi-based HAR by enabling transfer learning \cite{Fu234782379}.

\section{Related Work}
A comprehensive survey on the state of cross-domain WiFi sensing is presented by Chen et al. \cite{Chen2023}, discussing domain-invariant feature extraction, virtual sample generation, transfer learning, few-shot learning, and big data approaches, as well as open challenges limiting practical applicability. According to this classification scheme, this work falls into the category of virtual sample generation, using data augmentation to facilitate generalization. Only a few works investigate data augmentation techniques, with a primary focus on RSSI data augmentation \cite{Rizk2019, Liu2020, hilal2021dataloc}. In \cite{Gao9729785}, a data augmentation technique based on noise injection into CSI data is proposed. Lee et al. \cite{Lee10185958} are among the first to address the problem of long-term generalization in CSI-based HAR and propose a temporal-invariant feature extraction approach using MixUp \cite{zhang2018mixup} augmentation to generalize their model. Key findings of this work include that CSI temporal variation can occur even in static environments, movements of small reflective objects can significantly alter CSI, and that CSI temporal variation occurs differently depending on the subcarrier index and receiver location, highlighting the difficulty of the underlying domain generalization problem. Moreover, in a recent work \cite{serbetci2023simple}, simple data augmentation techniques for CSI phase and amplitude data are proposed.

Building on this work, we extend the investigated domains to cross-scenario (LOS and through-wall (NLOS)) and cross-system settings, utilizing two different antenna configurations, which remain under-explored.

\section{Experimental Setup}
We will now describe the experimental setup, including the hardware components of our systems, the characteristics of the test environment, and the protocol for gathering CSI amplitude activity spectrograms used for training CNN-based activity recognition models.

\begin{figure}[t!]
  \centering
  \begin{subfigure}{0.47\linewidth}
    \includegraphics[width=1.0\linewidth]{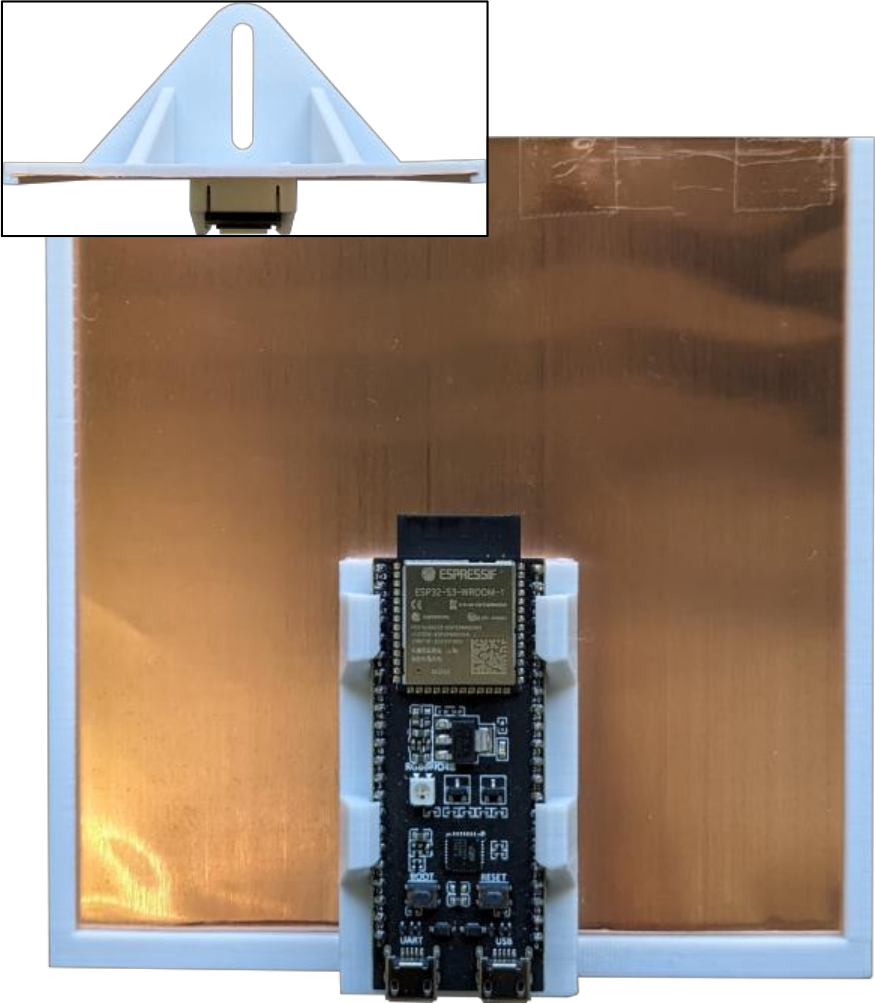}
    \caption{}
    \label{fig:antennasPIFA}
  \end{subfigure}
  \hfill
  \begin{subfigure}{0.51\linewidth}
    \includegraphics[width=1.0\linewidth]{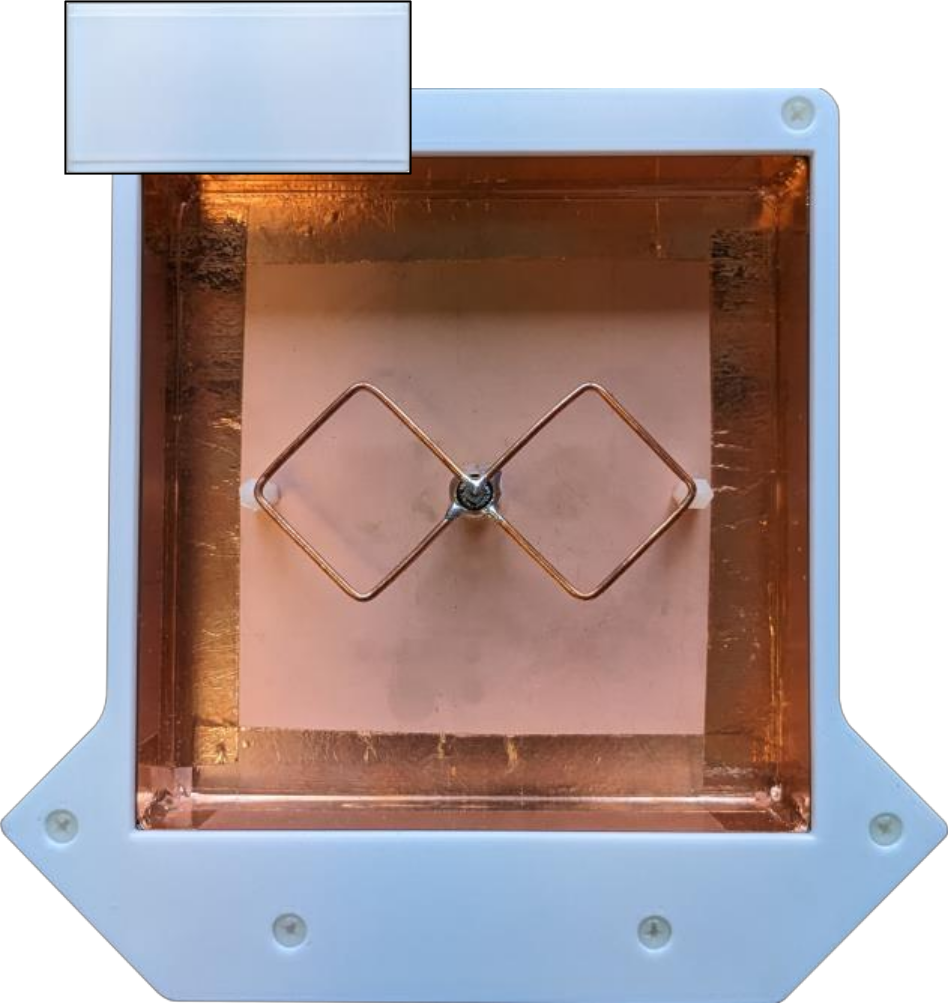}
    \caption{}
    \label{fig:antennasBiquad}
  \end{subfigure}
  \vspace{-5mm}
  \caption{Systems overview, showing (a) the PIFA with a plane reflector system, and (b) the BQ antenna system.}
  \label{fig:systemsOverview}
  \vspace{-6mm}
\end{figure}

\vspace{-1mm}
\subsection{Hardware}
We consider the two systems shown in Figure \ref{fig:systemsOverview}, which are built upon the ESP32-S3-DevKitC-1\footnote{ESP32-S3-DevKit-1, \href{https://docs.espressif.com/projects/esp-idf/en/latest/esp32s3/hw-reference/esp32s3/user-guide-devkitc-1.html}{https://docs.espressif.com}, accessed: 10-09-2023} development board, featuring an ESP32-S3-WROOM-1\footnote{ESP32-S3-WROOM-1, \href{https://www.espressif.com/sites/default/files/documentation/esp32-s3-wroom-1_wroom-1u_datasheet_en.pdf}{https://docs.espressif.com}, accessed: 10-09-2023} module for WiFi connectivity. The systems are deployed in a point-to-point transmitter-receiver configuration, where one of the two identical devices serves as a transmitter, sending CSI packets at a fixed frequency of 100Hz, while the other device functions as a receiver, continually listening for CSI packets. A WiFi connection between the transmitter and receiver is established using Espressif's wireless communication protocol, ESP-NOW\footnote{ESP-NOW, \href{https://www.espressif.com/en/solutions/low-power-solutions/esp-now}{https://docs.espressif.com}, accessed: 10-09-2023}, and CSI packets are captured using Espressif's IoT Development Framework, ESP-IDF\footnote{ESP-IDF, \href{https://docs.espressif.com/projects/esp-idf/en/latest/esp32/get-started/}{https://docs.espressif.com}, accessed: 10-09-2023}. 

\textbf{PIFA system.} The system depicted in Figure \ref{fig:antennasPIFA} utilizes the built-in meandered printed inverted-F antenna (PIFA)\footnote{PIFA, \href{https://www.ti.com/lit/an/swra117d/swra117d.pdf}{https://www.ti.com}, accessed: 10-09-2023} \cite{pradhan2013parametric} of the ESP32-S3-WROOM-1 module. The PIFA alone can provide basic WiFi connectivity in most traditional scenarios, however, it is not ideal for HAR applications. Its omnidirectionality not only prevents the constraining of the recording environment but also makes it susceptible to noise from outside the recording environment (e.g., a person walking behind the system or on the floor below) \cite{Hernandez2021}. Moreover, its low gain of 2 dBi could hinder the establishment of a stable connection in long-range through-wall HAR scenarios. To address these shortcomings without replacing the PIFA, we employ a plane reflector made from a 123$\times$123 mm, 0.2 mm thick copper sheet. Both the ESP32-S3-DevKitC-1 development board and the reflector are rigidly mounted to a 3D-printed frame, creating a 1/8-wavelength spacing of 15mm (2.4GHz carrier frequency) between the PIFA and the reflector. The added plane reflector eliminates noise originating from the backside of the PIFA and simultaneously increases its forward gain.

\begin{figure}[t!]
  \centering
  \includegraphics[width=\linewidth]{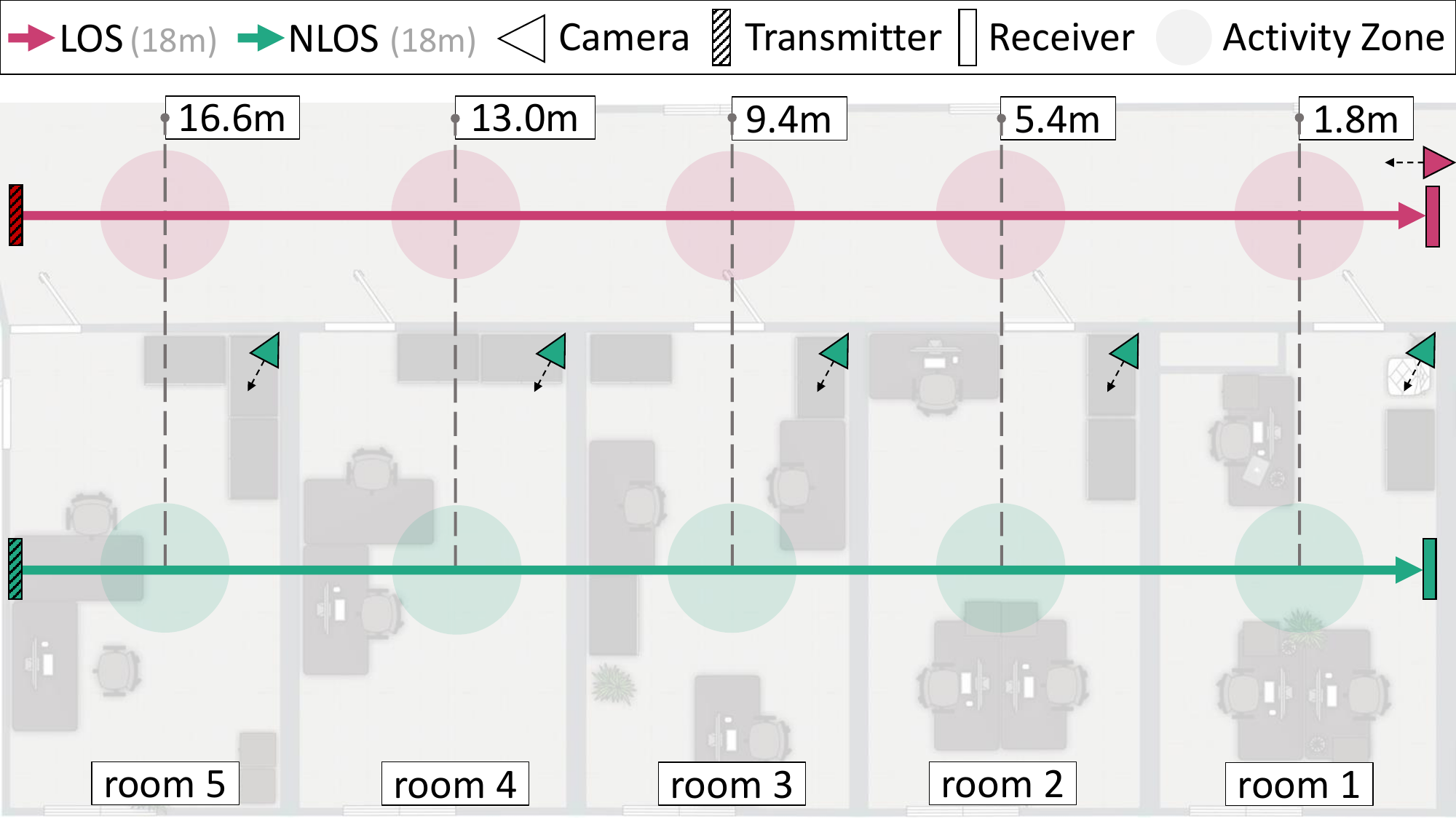}
  \caption{Floor plan of the test environment, showing the transmitter and receiver placement in LOS and NLOS scenarios.}
  \label{environment}
  \vspace{-6mm}
\end{figure}

\begin{figure*}[t!]
  \centering
  \begin{subfigure}{0.495\linewidth}
    \includegraphics[width=1.0\linewidth]{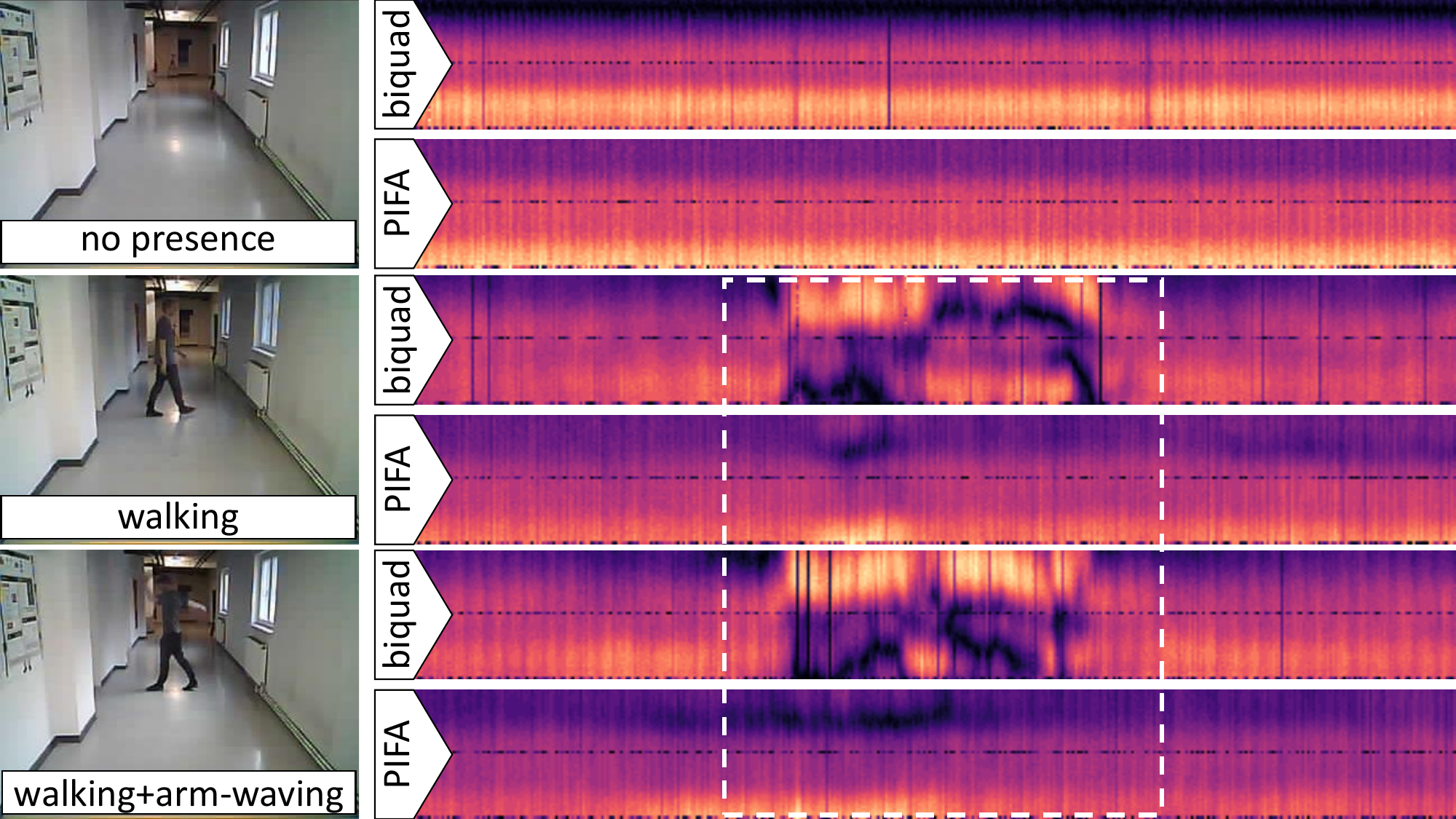}
    \caption{LOS CSI activity spectrograms from W1.8k.}
    \label{fig:LOS_activities}
    \vspace{2mm}
  \end{subfigure}
\hfill
  \begin{subfigure}{0.495\linewidth}
    \includegraphics[width=1.0\linewidth]{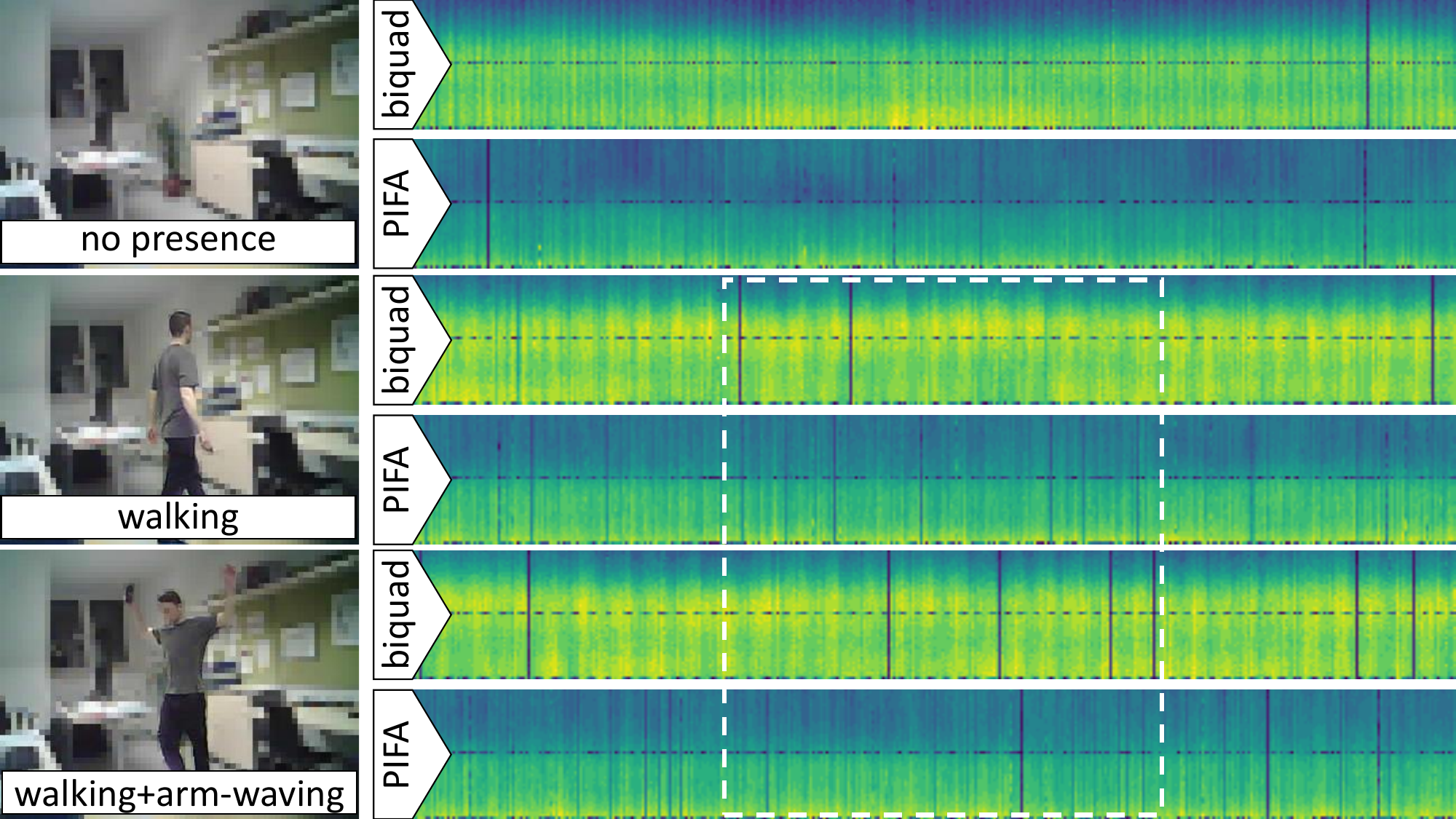}
    \caption{NLOS CSI activity spectrograms from W1.8k.}
    \label{fig:NLOS_activities}
    \vspace{2mm}
  \end{subfigure}
  \hfill
  \vspace{-5mm}
  \caption{(a) LOS and (b) NLOS CSI amplitude spectrograms of the classes \textit{no presence}, \textit{walking}, and \textit{walking + arm-waving}, captured with BQ antenna and PIFA with plane reflector systems at a distance of 9.4m (room 3 in the NLOS scenario). The spectrograms show the amplitudes of 52 L-LTF subcarriers over a time interval of 4 seconds ($\sim$400 packets).}
  \label{fig:activities}
  \vspace{-3mm}
\end{figure*}

\textbf{BQ system.} Our second system, shown in Figure \ref{fig:antennasBiquad}, replaces the PIFA with an external antenna. We chose a directional biquad (BQ) antenna design with a gain of 10-12 dBi and a beamwidth of 70$^{\circ}$ \cite{Singh2012ANB}. This choice strikes a balance between gain, compactness, and ease of construction using common materials. Moreover, while antennas with higher gain and extremely narrow beam widths exist for establishing long-range point-to-point connections, we favor a beamwidth around 70$^{\circ}$ for HAR applications, as it facilitates comprehensive room coverage in most scenarios while maintaining constraints on the recording environment. Additionally, its similarity to the field of view (FOV) of typical cameras allows easy integration with a camera having a corresponding FOV, providing a sense of the antenna beam's coverage area.

\subsection{Environment}
The test environment for our evaluation is depicted in Figure \ref{environment}. This environment comprises an 18m-long hallway connected to five adjacent rooms containing office furniture. These rooms, separated by 25cm thick brick walls, present a challenging long-range NLOS scenario. Moreover, the rooms are of uniform size (approximately 3.5m $\times$ 6.0m) and arranged in a manner that facilitates a direct comparison of LOS and NLOS HAR performance at various distances between the transmitter and receiver.

For the LOS scenario (red line), the transmitter and receiver are positioned at opposite ends of the hallway, facing each other. To capture activity images required for the annotation of raw CSI data, an additional ESP32-S3-based camera board is placed next to the transmitter, aligned in its direction.

In the NLOS scenario (green line), the transmitter and receiver once again face each other but are placed at the outer walls of rooms 5 and 1, respectively. The alignment of antennas is achieved by fine-adjusting the receiver's horizontal position in the room based on the RSSI at the receiver's end. As in the LOS scenario, activity images are captured in the NLOS scenario using an ESP32-S3-based camera board placed in the room where the activity occurs. A constant transmitter-receiver spacing of 18m is maintained throughout all experiments for both LOS and NLOS scenarios.

\subsection{Data}
To assess cross-scenario (LOS$\leftrightarrows$NLOS) and cross-system (PIFA$\leftrightarrows$BQ) HAR performance, we collect and publicly make available the Wallhack1.8k dataset\footnote{Wallhack1.8k, \href{https://zenodo.org/record/8188999}{https://zenodo.org/record/8188999}}, which comprises 1806 CSI amplitude spectrograms of human activities collected in the test environment. We utilize this dataset for training CNN-based HAR models. The objective is to distinguish between coarse and fine body movements (e.g., walking vs. arm movements). In the data collection process, activities are conducted within five circular activity zones (1.5m radius) along the LOS and NLOS transmission paths, as shown in Figure \ref{environment}. These zones are located at distances of \{1.8, 5.4, 9.4, 13.0, 16.6\}m from the receiver, corresponding to room centers in the NLOS scenario. For both systems, we record two minutes of continuous \textit{walking} and \textit{walking + arm-waving} activities in each activity zone, as well as five minutes of \textit{no presence} (no person in the recording environment) for each scenario. 

As a pre-processing step, the recorded CSI time series data is trimmed using the corresponding RGB images to remove any CSI packets that do not correspond to the target activity. The trimmed CSI time series data is then transformed into time-frequency plots of subcarrier amplitudes over time, creating spectrograms. To achieve this, the trimmed CSI time series data is divided into segments of 400 CSI packets (equivalent to 4-second time intervals at a sending frequency of 100Hz), and the amplitudes of 52 Legacy Long Training Field (L-LTF) subcarriers are plotted. This results in a spectrogram size of 400$\times$52. 

\begin{table}[t!]
\centering
\caption{Distribution of samples across subsets of the Wallhack1.8k dataset. }
\label{wallhackOverview}
\begin{tabularx}{0.48\textwidth}{>{\arraybackslash}p{11mm}>{\centering\arraybackslash}p{11mm}>{\centering\arraybackslash}p{10mm}>{\centering\arraybackslash}p{10mm}>{\centering\arraybackslash}p{10mm}Y}
\toprule
Subset & Scenario & System & Rooms & Activities & Samples\\
\midrule
W1.8k$_{LB}$ & LOS & BQ & 1 & 3 & 458\\
W1.8k$_{LP}$ & LOS & PIFA  & 1 & 3 & 461\\
W1.8k$_{NB}$ & NLOS & BQ  & 5 & 3 & 450\\
W1.8k$_{NP}$ & NLOS & PIFA & 5 & 3& 437\\
\midrule    
 &  &  &  & Total: & 1806\\
\bottomrule
\end{tabularx}
\vspace{-2mm}
\end{table}

\begin{table}[t!]
\caption{Distribution of activity classes within the subsets of Wallhack1.8k.}
\centering
\begin{tabularx}{0.48\textwidth}{>{\arraybackslash}p{11mm}>{\centering\arraybackslash}p{16mm}>{\centering\arraybackslash}p{14mm}>{\centering\arraybackslash}p{32mm}}
\toprule
Subset & \textit{no presence}  & \textit{walking} &  \textit{walking + arm-waving}  \\
\midrule
W1.8k$_{LB}$ & 149 & 154 & 155\\
W1.8k$_{LP}$ & 149 & 160 & 152\\
W1.8k$_{NB}$ & 148 & 150 & 152\\
W1.8k$_{NP}$ & 143 & 147 & 147\\
\midrule    
 Total: & 589 & 611 & 606\\
\bottomrule
\end{tabularx}
\label{wallhackClasses}
\vspace{-2mm}
\end{table}

For the supervised training of activity recognition models, spectrograms are assigned the labels \{0, 1, 2\}, corresponding to the classes \textit{no presence}, \textit{walking}, and \textit{walking+arm-waving}. This data collection process for both systems in both LOS and NLOS scenarios leads to four subsets constituting the Wallhack1.8k dataset. We adopt the subset naming convention W1.8k$_{XY}$, where X represents the scenario (L=LOS or N=NLOS) and Y represents the system (B=BQ or P=PIFA). The detailed distributions of the 1806 spectrograms in Wallhack1.8k across subsets and classes are provided in Tables \ref{wallhackOverview} and \ref{wallhackClasses}, respectively. Moreover, Figure \ref{fig:activities} shows characteristic spectrogram examples for all systems, scenarios, and classes. Upon visually examining the LOS spectrograms in Figure \ref{fig:LOS_activities}, the pronounced amplitude patterns in spectrograms collected by the BQ system are evident. In comparison, the amplitude patterns in spectrograms collected by the PIFA system are hard to discern from the WiFi background. This suggests that the BQ system exhibits higher sensitivity to human activities. In the NLOS scenario, depicted in Figure \ref{fig:NLOS_activities}, no activity-specific amplitude patterns are visible in the spectrograms of either system. These observations show the high sensitivity of CSI to environmental conditions and sensing hardware, resulting in drastically different data characteristics.

\begin{table}[t!]
\caption{Cross-scenario generalization performance of activity recognition models, trained on spectrograms collected with the PIFA system. X$\rightarrow$Y indicates model training on data from scenario X and testing on data from scenario Y.}
\centering
\begin{tabularx}{0.475\textwidth}{>{\arraybackslash}p{27mm}>{\arraybackslash}p{11mm}>{\arraybackslash}p{9mm}>{\arraybackslash}p{11mm}>{\arraybackslash}p{9mm}}
\toprule
Augmentation & LOS$\rightarrow$NLOS &  & NLOS$\rightarrow$LOS & \\
\midrule
\textit{none} & 37.5$\pm$5.3 &  & 65.0$\pm$5.5 &\\
\textit{randomCircularRotation} & 43.6$\pm$9.8 & $\uparrow$ 6.1 & 61.3$\pm$6.4 & $\downarrow$ 3.7\\
\textit{randomResizedCrop} & 49.5$\pm$11.4 & $\uparrow$ 12.0 & 54.8$\pm$8.4 & $\downarrow$ 10.2\\
\textit{randomAmplitude} & 34.1$\pm$7.5 & $\downarrow$ 3.4 & 66.1$\pm$4.8  & $\uparrow$ 1.1\\
\textit{randomContrast} & 32.5$\pm$3.2 & $\downarrow$ 5.0 & 63.9$\pm$5.9 & $\downarrow$ 1.1\\
\midrule
\textit{combined} & 36.4$\pm$7.1 & $\downarrow$ 1.1 & 66.1$\pm$4.8  & $\uparrow$ 1.1\\
\bottomrule
\end{tabularx}
\label{crossScenarioPIFA}
\vspace{-3mm}
\end{table}

\subsection{Augmentations}
The four investigated data augmentations in our ablation study are \textit{randomCircularRotation}, \textit{randomResizedCrop}, \textit{randomAmplitude}, and \textit{randomContrast}, which are applied to CSI amplitude spectrograms with a probability of $p=0.5$.

\textbf{\textit{randomCircularRotation.}} Treating the spectrogram as a 2D time-frequency array, circular rotations along the time axis shift the array elements along the positive time direction, with elements going beyond the array's boundary wrapping around to the opposite side, becoming the first element. A random number of circular rotations $n \sim \mathcal{U}(1, w)$, with $w=400$ being the spectrogram width, is applied to the spectrogram.

\textbf{\textit{randomResizedCrop.}} For this augmentation, the spectrogram is either cropped along the time axis and stretched back to its original width $w$ or compressed along the time axis and re-scaled to $w$. Both the crop and the compression factor are randomly sampled from $\mathcal{U}(\frac{w}{2}, w)$. The cropping and re-scaling of the spectrogram correspond to a slowdown in time, while the compression and re-scaling correspond to a speed-up, resulting in new samples with activities being carried out at varying speeds.

\textbf{\textit{randomAmplitude.}} As observed in \cite{Lee10185958}, CSI amplitude can vary significantly even in static environments and system settings. To replicate this behavior, we randomly scale the amplitude of spectrograms on a per-channel basis. The magnitude of the augmentation factor is randomly sampled from $\mathcal{U}(0.75, 1.25)$.

\textbf{\textit{randomContrast.}} Furthermore, it is also found that CSI amplitude variation occurs differently depending on the subcarrier index \cite{Lee10185958}. Following the approach of the \textit{randomContrast} augmentation, this behavior is replicated by randomly scaling the contrast of spectrograms on a per-channel basis. The magnitude of the augmentation factor is randomly sampled from $\mathcal{U}(0.75, 1.25)$.

\subsection{Model Training}
As a platform for evaluating cross-scenario and cross-system generalization, we use the EfficientNetV2 architecture \cite{tan2021efficientnetv2}, utilizing the standard implementation from \emph{torchvision.models}\footnote{EfficientnetV2s, \href{https://pytorch.org/vision/main/models/efficientnetv2.html}{https://pytorch.org/}, accessed: 10-09-2023}. EfficientNetV2 small is a lightweight feature extractor commonly employed as a backbone. All models are trained from scratch to eliminate any prior knowledge derived from pre-training on (ImageNet) RGB images that could influence the results. The models undergo 400 epochs of training using the Adam optimizer with a learning rate of 0.0001 and a batch size of 16. Additionally, a balanced sampler is used to mitigate class imbalances in the training dataset. For each model, we conduct ten independent training runs and select the model instance with the highest validation accuracy from each run. To acquire representative results, the mean and standard deviation of classification accuracies across ten runs are employed as evaluation metrics.

\begin{table}[t!]
\caption{Cross-scenario generalization performance of activity recognition models, trained on spectrograms collected with the BQ system. X$\rightarrow$Y indicates model training on data from scenario X and testing on data from scenario Y.}
\centering
\begin{tabularx}{0.475\textwidth}{>{\arraybackslash}p{27mm}>{\arraybackslash}p{11mm}>{\arraybackslash}p{9mm}>{\arraybackslash}p{11mm}>{\arraybackslash}p{9mm}}
\toprule
Augmentation & LOS$\rightarrow$NLOS & & NLOS$\rightarrow$LOS & \\
\midrule
\textit{none} & 36.4$\pm$5.1 & & 53.0$\pm$10.1 &\\
\textit{randomCircularRotation} & 39.6$\pm$9.6 & $\uparrow$ 3.2 & 69.6$\pm$3.2 & $\uparrow$ 16.6\\
\textit{randomResizedCrop} & 41.8$\pm$7.4 & $\uparrow$ 5.4 & 63.3$\pm$6.9 & $\uparrow$ 10.3\\
\textit{randomAmplitude} & 41.1$\pm$5.2 & $\uparrow$ 4.7 & 61.7$\pm$6.2  & $\uparrow$ 8.7\\
\textit{randomContrast} & 40.2$\pm$5.3 & $\uparrow$ 3.8 & 57.0$\pm$8.1  & $\uparrow$ 4.0 \\
\midrule
\textit{combined} & 58.9$\pm$7.7 & $\uparrow$ 22.5 & 68.7$\pm$2.9 & $\uparrow$ 15.7\\
\bottomrule
\end{tabularx}
\label{crossScenarioBQ}
\vspace{-3mm}
\end{table}

\section{Ablation Study}
To assess the effects of data augmentations on cross-scenario (LOS$\leftrightarrows$NLOS) and cross-system (PIFA$\leftrightarrows$BQ) generalization performance of our models, an ablation study is conducted. Starting with a baseline model trained without data augmentation, we train separate models with each data augmentation and measure the change in activity recognition accuracy. Beneficial data augmentations that lead to an increase in accuracy are combined for the training of a final model, being also compared against the baseline.

\subsection{Cross-scenario Generalization}
\textbf{PIFA system.} The ablation study results, utilizing data captured by the PIFA system, are presented in Table \ref{crossScenarioPIFA}. When considering generalization from the LOS to the NLOS scenario (LOS$\rightarrow$NLOS), we see improvements over the baseline performance of 37.5$\pm$5.3 by employing the \textit{randomCircularRotation} and \textit{randomResizedCrop} augmentations. These augmentations result in an increase in model accuracy by 6.1 and 12.0 percentage points, respectively. Interestingly, when combining both augmentations, a decrease in model accuracy by 1.1 percentage points is observed, suggesting a complex relationship between these augmentations. In the reverse scenario (NLOS$\rightarrow$LOS), we see a notably higher baseline performance of 65.0$\pm$5.5, indicative of asymmetric generalization between scenarios. Furthermore, the augmentation that demonstrates an improvement of 1.1 percentage points over the baseline is solely the \textit{randomAmplitude} augmentation.

\textbf{BQ system.} The ablation study results based on data captured by the BQ system are presented in Table \ref{crossScenarioBQ}. For the generalization from the LOS to the NLOS scenario (LOS$\rightarrow$NLOS), we observe moderate improvements over the baseline performance of 36.4$\pm$5.1 for all data augmentations. Furthermore, training with all data augmentations leads to a significant improvement in model accuracy of 22.5 percentage points. Also, for the inverse direction (NLOS$\rightarrow$LOS), all data augmentations have a positive effect on model accuracy. Especially the \textit{randomCircularRotation} augmentation leads to a significant improvement of 16.6 percentage points. Interestingly, although individual data augmentations all improve model accuracy, training with a combination of all augmentations slightly reduces model performance, leading to a smaller improvement over baseline accuracy of 15.7 percentage points.

\begin{table}[t!]
\caption{Cross-system generalization performance of activity recognition models, trained on LOS spectrograms. X$\rightarrow$Y indicates model training on data from system X and testing on data from system Y.}
\centering
\begin{tabularx}{0.475\textwidth}{>{\arraybackslash}p{27mm}>{\arraybackslash}p{11mm}>{\arraybackslash}p{9mm}>{\arraybackslash}p{11mm}>{\arraybackslash}p{9mm}}
\toprule
Augmentation & PIFA$\rightarrow$BQ & & BQ$\rightarrow$PIFA &\\
\midrule
\textit{none} & 36.1$\pm$6.0 & & 35.2$\pm$4.7 &\\
\textit{randomCircularRotation} & 48.7$\pm$13.5 & $\uparrow$ 12.6 & 38.9$\pm$2.5 & $\uparrow$ 3.7\\
\textit{randomResizedCrop} & 38.7$\pm$6.7 & $\uparrow$ 2.6 & 36.5$\pm$3.0 & $\uparrow$ 1.3\\
\textit{randomAmplitude} & 36.5$\pm$4.1 & $\uparrow$ 0.4 & 35.2$\pm$4.9 & $\sim$0.0\\
\textit{randomContrast} & 36.5$\pm$4.8 & $\uparrow$ 0.4 & 32.4$\pm$3.9 & $\downarrow$ 2.8\\
\midrule
\textit{combined} & 57.2$\pm$15.7 & $\uparrow$ 21.1 & 50.7$\pm$11.8 & $\uparrow$ 18.5\\
\bottomrule
\end{tabularx}
\label{crossSystemLOS}
\vspace{-3mm}
\end{table}

\subsection{Cross-system Generalization}
\textbf{LOS scenario.} We provide ablation study results for cross-system generalization based on LOS data in Table \ref{crossSystemLOS}. When transitioning from the PIFA system to the BQ system (PIFA$\rightarrow$BQ), we observe improvements over the baseline performance of 36.1$\pm$6.0 across all data augmentations. Notably, the \textit{randomCircularRotation} augmentation significantly enhances model accuracy by 12.6 percentage points. Furthermore, combining all data augmentations results in a 21.1 percentage point improvement compared to the baseline. In the reverse scenario (BQ$\rightarrow$PIFA), we observe a similar trend. While individual data augmentations yield moderate improvements, training with all augmentations leads to a significant model accuracy improvement, surpassing the baseline by 18.5 percentage points.

\textbf{NLOS scenario.} Ablation study results for cross-system generalization based on NLOS data are presented in Table \ref{crossSystemNLOS}. When transitioning from the PIFA system to the BQ system (PIFA$\rightarrow$BQ), improvements over the baseline performance of 34.7$\pm$6.5 are observed with all data augmentations except \textit{randomContrast}. Combining all augmentations with a positive effect on model accuracy does not yield additional improvements, achieving a 4.9 percentage point gain over the baseline (the same as with \textit{randomCircularRotation}). A clearer trend is observable when transitioning from the BQ system to the PIFA system (BQ$\rightarrow$PIFA), where all data augmentations result in a moderate improvement in model accuracy over the baseline of 30.0$\pm$3.9. Furthermore, the combination of all augmentations results in a significant improvement of 22.3 percentage points over the baseline.

\begin{table}[t!]
\caption{Cross-system generalization performance of activity recognition models, trained on NLOS spectrograms. X$\rightarrow$Y indicates model training on data from system X and testing on data from system Y.}
\centering
\begin{tabularx}{0.475\textwidth}{>{\arraybackslash}p{27mm}>{\arraybackslash}p{11mm}>{\arraybackslash}p{9mm}>{\arraybackslash}p{11mm}>{\arraybackslash}p{9mm}}
\toprule
Augmentation & PIFA$\rightarrow$BQ & & BQ$\rightarrow$PIFA & \\
\midrule
\textit{none} & 34.7$\pm$6.5 & & 30.0$\pm$3.9 &\\
\textit{randomCircularRotation} & 39.6$\pm$6.8 & $\uparrow$ 4.9 & 34.5$\pm$5.7 & $\uparrow$ 4.5\\
\textit{randomResizedCrop} & 39.3$\pm$3.2 & $\uparrow$ 4.3 & 31.4$\pm$10.5 & $\uparrow$ 1.4\\
\textit{randomAmplitude} & 36.4$\pm$6.1 & $\uparrow$ 1.7 & 33.0$\pm$4.8 & $\uparrow$ 3.0\\
\textit{randomContrast} & 33.8$\pm$6.6 & $\downarrow$ 0.9 & 33.6$\pm$4.4 & $\uparrow$ 3.6\\
\midrule
\textit{combined} & 39.6$\pm$8.2 & $\uparrow$ 4.9 & 52.3$\pm$13.8 & $\uparrow$ 22.3\\
\bottomrule
\end{tabularx}
\label{crossSystemNLOS}
\vspace{-3mm}
\end{table}

\section{\uppercase{Discussion}}
The results in Tables \ref{crossScenarioPIFA} and \ref{crossScenarioBQ} reveal an asymmetric relationship in cross-scenario generalization. For both systems, models generalize better from NLOS to LOS scenarios than the other way around. This is indicated by the relatively high baseline accuracies, which are well above random guessing performance ($\frac{1}{3}$).

Another observation, as indicated by the results in Table \ref{crossScenarioBQ}, is that the BQ system is more receptive to data augmentations, indicating that the domain gap between scenarios is narrower for this system. A possible explanation is that the higher gain and directionality make the system less susceptible to environmental obstructions in the transmission path, and consequently, the data characteristics of both scenarios are more similar.

Furthermore, the results in Tables \ref{crossSystemLOS} and \ref{crossSystemNLOS} show a smaller domain gap between systems in the LOS scenario, as indicated by the similar baseline accuracies and high receptiveness to data augmentation in the LOS scenario. For the NLOS scenario, we see an asymmetric relationship, with models generalizing better from the BQ system to the PIFA system than the other way around (22.3 vs. 4.9 percentage points).

Finally, our ablation study shows that cross-scenario and cross-system generalization via data augmentation do require careful selection of data augmentation techniques, as for each case, the individual and combined effects of data augmentation can vary significantly.

\section{\uppercase{Conclusion}}
In this work, we investigated the effectiveness of four data augmentation techniques commonly applied to images on cross-scenario and cross-system model performance in WiFi CSI-based HAR. We collected a dataset of LOS and NLOS CSI spectrograms captured by two different WiFi systems and used it to train HAR models in an ablation study, examining the effects on model performance. The gathered results show that specific combinations of the investigated data augmentation techniques can significantly improve both cross-scenario and cross-system performance.

\bibliographystyle{IEEEtran}
\bibliography{IEEEexample}

\end{document}